\documentclass[conference,a4paper]{IEEEtran}
\IEEEoverridecommandlockouts
\usepackage{cite}
\usepackage{amsmath,amssymb,amsfonts}
\usepackage{algorithmic}
\usepackage{graphicx}
\usepackage{textcomp}
\usepackage{siunitx}
\usepackage{balance}
\usepackage{subcaption}
\usepackage{caption}
\def\BibTeX{{\rm B\kern-.05em{\sc i\kern-.025em b}\kern-.08em
    T\kern-.1667em\lower.7ex\hbox{E}\kern-.125emX}}

\usepackage{siunitx}
\usepackage{hyperref}
\hypersetup{
    colorlinks=true,
    linkcolor=black,
    filecolor=magenta,      
    urlcolor=cyan,
    citecolor=black
}
\usepackage[table]{xcolor}

\begin{document}

\title{Open-Source LiDAR Time Synchronization System by Mimicking GNSS-clock
}

\author{\IEEEauthorblockN{Marsel Faizullin\IEEEauthorrefmark{1}, Anastasiia Kornilova\IEEEauthorrefmark{2} and Gonzalo Ferrer\IEEEauthorrefmark{3}}
\IEEEauthorblockA{
Skolkovo Institute of Science and Technology, Moscow, Russia \\
\textit{ORCID: \IEEEauthorrefmark{1}0000-0002-1053-7771, \IEEEauthorrefmark{2}0000-0002-2267-9689, \IEEEauthorrefmark{3}0000-0003-2704-7186}}
}

\maketitle

\begin{abstract}
Data fusion algorithms that employ LiDAR measurements, such as Visual-LiDAR, LiDAR-Inertial, or Multiple LiDAR Odometry and simultaneous localization and mapping (SLAM) rely on precise timestamping schemes that grant synchronicity to data from LiDAR and other sensors.
Poor synchronization performance, due to incorrect timestamping procedure, may negatively affect the algorithms' state estimation results.
To provide highly accurate and precise synchronization between the sensors, we introduce an open-source hardware-software LiDAR to other sensors time synchronization system that exploits a dedicated hardware LiDAR time synchronization interface by providing emulated GNSS-clock to this interface, no physical GNSS-receiver is needed. 
The emulator is based on a general-purpose microcontroller and, due to concise hardware and software architecture, can be easily modified or extended for synchronization of sets of different sensors such as cameras, inertial measurement units (IMUs), wheel encoders, other LiDARs, etc.
In the paper, we provide an example of such a system with synchronized LiDAR and IMU sensors.
We conducted an evaluation of the sensors synchronization accuracy and precision, and state 1 \si{\micro\second} performance. 
We compared our results with timestamping provided by ROS software and by a LiDAR inner clocking scheme to underline clear advantages over these two baseline methods.







\end{abstract}


\section{Introduction}


\begin{figure}[t]
    \begin{subfigure}{\columnwidth}
        \includegraphics[width=\columnwidth]{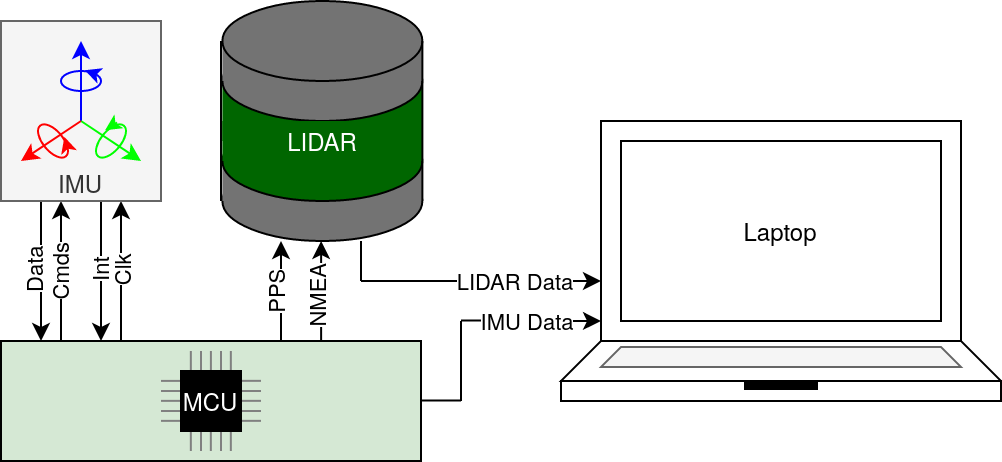}
        \caption{Block-scheme of the system}
        \label{fig_blocksub}
    \end{subfigure}
    \vskip\baselineskip
    \begin{subfigure}{\columnwidth}
        \includegraphics[width=\columnwidth]{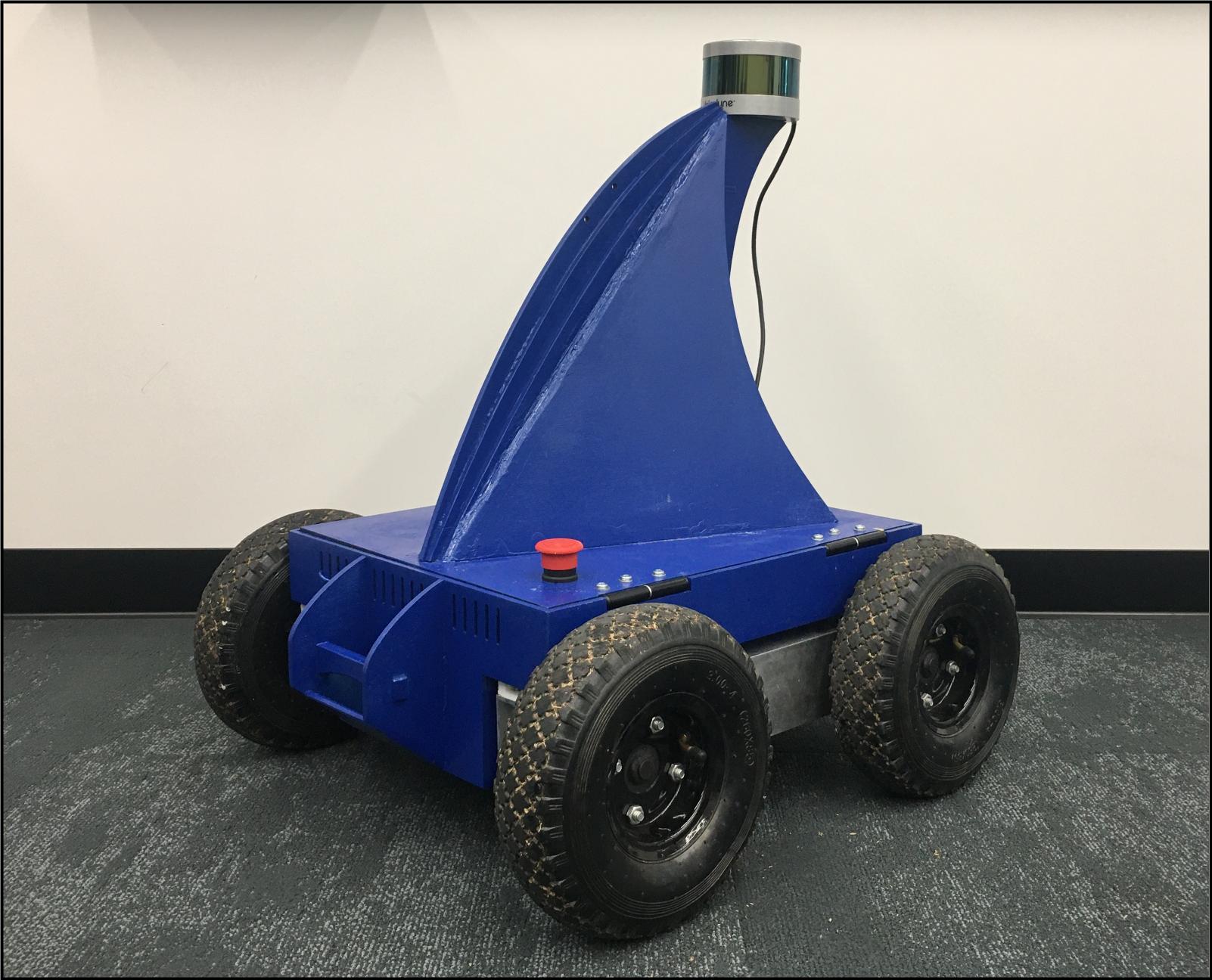}
        \caption{Common view of the moving platform}
        \label{fig_common}
    \end{subfigure}
    \caption{Block-scheme of the system (a) and common view of a 4WD moving platform carrying the system (b). \textit{Data} and \textit{Cmds} lines are used for IMU data gathering and IMU setup, \textit{Int} is the interrupt line for precise timestamping, \textit{Clk} line -- reference clock. \textit{PPS} -- PPS signals, \textit{NMEA} -- NMEA GPRMC messages. Signal directions are depicted by arrows. IMU, MCU-board, and laptop, along with miscellaneous equipment of the platform, are placed into the platform body. Additional information on the system can be found at \url{https://github.com/MobileRoboticsSkoltech/lidar-sync-mimics-GPS} or \url{https://github.com/Marselka/lidar-sync-mimics-GPS}.}
    \label{fig_block}
    \vspace{-0.7cm}
\end{figure}








Time synchronization (or shortly sync) is a crucial stage in the development of sensor networks aiming to solve data fusion, robot navigation or simultaneous localization and mapping (SLAM) problems. Synchronization matches measurements of different sensors with different clocks and thus allows a sensor network to create a common timing reference recorded by different clocks or systems.


Neglecting synchronization considerations generates conflicting data that are hard to fuse by the algorithms.
A time offset of several \SI{}{\ms} between measurements of two LiDARs on high dynamic conditions introduces a significant position error of the same observed object. For instance, during a multi-LiDAR platform rotation on a spot (differential wheeled robot) with 5 radian per second angular velocity, time offset of 10 milliseconds leads to rotation error of 3 degrees that corresponds to 50 cm shift between observations of an object located at 10 meter distance. Other examples that underline essentiality of sensor synchronization are considered in~\cite{olson2010passive, english2015triggersync, nikolic2014synchronized, tschopp2020versavis}.


Time synchronization methods may be divided into hardware (HW) and software (SW) methods. 
Precise \textit{hardware} synchronization based on GNSS-supplied clock is commonly used for sensors that have a specific interface for obtaining time from GNSS-receivers~\cite{velodyne}. However, a poor GNSS signal in GNSS-denied zones or urban canyon negatively affects not only localization but synchronization accuracy and precision as well~\cite{dana1997global}. 

A wide number of projects in the field of robotics are based on Robot Operation System (ROS) software\cite{quigley2009ros}. This software by default utilizes one of the simplest methods of \textit{software} time sync.
Timestamping of sensor measurements in ROS is usually performed by matching the time of arrival of messages to a ROS-powered computer. However, the accuracy and precision of this technique are far from desired. 
The characteristic fluctuation of time of arrived messages usually reaches several milliseconds and more while delays between the true moment of measurements vary from one interface to another~\cite{park2020real}.

In this work, we introduce hardware LiDAR to inertial measurement unit (IMU) time synchronization system and provide estimation of its accuracy and precision. Any other sensor can be installed in place of IMU while keeping precision value at the same level.
The system combines advantages of precise HW synchronization and ubiquitous open-source ROS software. At the same time, the system is entirely independent of the computer and can be used without ROS. 
Moreover, it does not need any specific synchronization equipment such as GNSS-receivers, and is based on general-purpose microcontroller (microcontroller unit or MCU).

The system is highly extendable and can manage synchronization of any other additional sensor by simple modification of firmware program blocks. It efficiently manages HW resources making it available for implementation even on a low-end MCU. We also provide synchronization precision analysis of our system, comparing the results with the pure ROS timestamping technique. Our system keeps the same magnitude of precision as the best possible on original separate timestamping schemes of the sensors. 
The current setup is used for correct data gathering for LiDAR-IMU Odometry algorithms~\cite{ye2019tightly}.
The hardware design, software, and firmware are open-source, and available at our project page~\cite{lidarimusync}. 

\section{Synchronization Principle and System Architecture}\label{sec_arch}



In this section we briefly describe the LiDAR synchronization principle. 
We have chosen the VLP-16 LiDAR~\cite{velodyne} due to its wide use in literature \cite{yeong2021sensor, shan2020lio, ponnaganti2020deep, sahin2021methods}.
The principle below is generalizable, and correct for any other LiDAR that supports connection to an external GNSS-receiver.
After the description, we introduce the HW architecture of our system. 

VLP-16 LiDAR possess a dedicated HW interface for synchronizing data with global time provided by a GNSS-receiver~\cite{velodyne}. 
The interface, through the physical connection, receives
\begin{itemize}
    \item pulse-per-second (PPS) electrical signal and
    \item NMEA GPRMC (NG) messages.
\end{itemize}
NMEA GPRMC stands for National Marine Electronics Association Global Positioning Recommended Minimum Specific GNSS Data, more information can be found in the respective standard~\cite{nmea}. 

\begin{figure}[t]
    \includegraphics[width=\columnwidth]{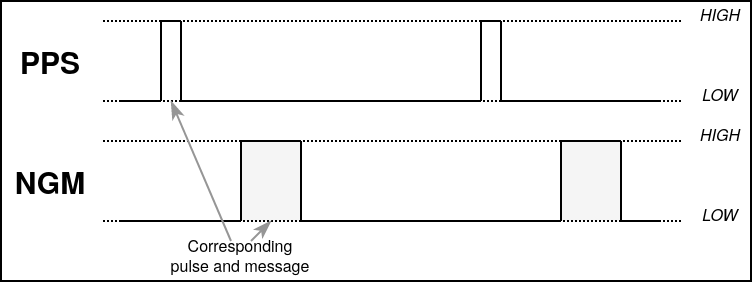}
    \caption{Output of a GNSS-receiver clock. Pulse-per-second (PPS) signals are followed by NMEA GPRMC messages (NGM). The signal and messages are mimicked by our system.}
\label{fig_pps}
\end{figure}

GNSS-receiver PPS signals consist of short pulses arising every 1 second, they provide HW synchronization between the global GNSS and the internal LiDAR clock. Every PPS signal is followed by its corresponding NG message (Fig.~\ref{fig_pps}). 
Every NG message is an ASCII string that contains a header, time, date, global coordinates, 
checksum and other information stored in fields separated by commas~\cite{gprmc}.
The LiDAR parses the string to obtain a timestamp. An example of an NG message is\\
\\
\texttt{\$GPRMC,\textbf{144326},A,5107.0017737,N,...}.\\
\\ 
The second field (bold) corresponds to GNSS-clock timestamp of the PPS signal, and in this example counts 14 hours 43 minutes 26 seconds of Coordinated Universal Time (UTC). The timestamp may also include sub-seconds. The date is also parsed but is not shown in the example. Thus, the combination of the pulses and messages allows the LiDAR to adjust its internal clock to the global clock. 
Our system emulates GNSS-receiver by the MCU-platform. 
We applied this principle to synchronize the LiDAR sensor on the HW level and developed a system that emulates GNSS-receiver by the MCU-platform and synchronizes LiDAR and IMU sensors. In our case, no global time is given, only local time synchronization between the sensors is applied. Below is the description of the system architecture.

The system consists of three main parts: 
\begin{enumerate}
    \item STM32F4DISCOVERY MCU platform, 
    \item VLP-16 LiDAR, 
    \item MPU-9150 IMU.
\end{enumerate}
All the data retrieved by the system are sent to a laptop or mini PC to be processed or stored: LiDAR directly to the laptop, IMU through MCU-platform. 

The MCU platform performs (i) GNSS-clock emulation and feeding the clock to LiDAR, (ii) timestamped IMU data gathering and transmission. Thus, the LiDAR already contains the MCU-based clock within its data packets. IMU samples are timestamped by the MCU as well. Other sensors may need a different synchronization and/or data gathering schemes and interfaces; however, the common clock emulation principles described in this work can be applied to them with no or minor modifications.

Therefore, the system timestamping authority is the MCU HW clock that guarantees sub-microsecond precise sync.
The block-scheme of the system along with the common view of the platform powered by the system are depicted in~Fig.~\ref{fig_block}. 

An IMU data interface cannot provide precise timings and, consequently, information on the time of every measurement. To reach a precise timestamping, we employ the IMU interrupt functionality~\cite{imu}. This tool provides an electrical pulse right after every new IMU data sample is ready. The application of this feature is explained in the next section. In addition, the IMU is provided with an external quartz crystal reference clock from MCU platform~\cite{schmid2008exploiting} to guarantee stability of the 1 kHz IMU sample rate. 


\section{Algorithm}\label{sec_algo}

\begin{figure}[t]
\includegraphics[width=\columnwidth]{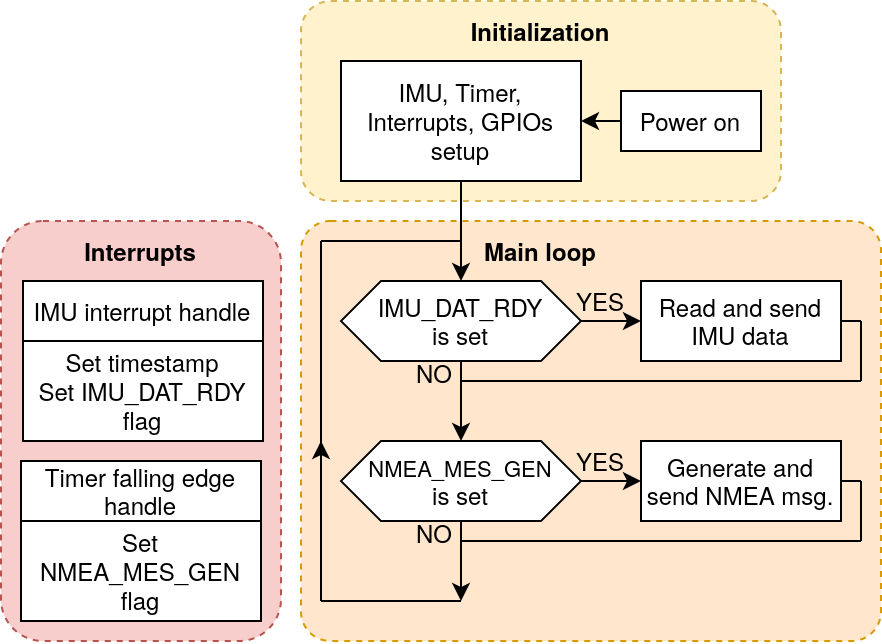}
\caption{The algorithm of the system. After initialization, the main loop is being interrupted by interrupts where specific flags are set. The flags are being checked in the main loop, and if some flag is set, specific operations are executed after the flag is reset.}
\label{fig_algo}
\end{figure}


The LiDAR-IMU synchronization and IMU data gathering algorithm consists of three main components depicted in Fig.~\ref{fig_algo}. The components are (i) initialization, (ii) main loop, (iii) interrupts. 

Initialization is aimed to set up all peripherals such as timer, the IMU reference clock, general-purpose input/output (GPIO), interrupts and other standard parameters and peripheral devices. The IMU is also set up at this stage. Although initialization is a common process, the role of the timer needs description in more details.

For emulation of a GNSS-receiver, the MCU-platform must (i) count time, (ii) generate PPS signals, and (iii) send NG messages to the LiDAR. Due to efficient firmware design, only a single general-purpose timer is utilized for performing these three processes. 
Otherwise, up to three timers would be needed.

For the first purpose (time count), the timer is used as a real-time clock (RTC) for IMU and LiDAR timestamping. The resolution of RTC is $1/76.8$ \si{\micro\second} (about 13 \SI{}{\ns}), such a high resolution allows for the assignment of timestamps of the measurements with high precision. The timer overflow happens at every 1 second; this trick allows the timer peripheral to generate a positive electrical pulse on an MCU GPIO-pin at every restart of the timer. This electrical pulse is nothing but the rising edge of a PPS signal for the LiDAR. 

The falling edge of the PPS signal, the negative pulse, is programmed to be at a half of a second after the positive one. This is the second usage of the timer. The negative pulse is also used to trigger an interrupt that sets up the NMEA\_MES\_GEN flag and makes the MCU generate and send an NG message containing a new timestamp to LiDAR; this is the third use of the timer. The exact time of sending the NG message is not critical for synchronization performance but must follow requirements described in~\cite{velodyne}. NG messages do not include sub-second information since the PPS rising edges happens exactly at every new second with zero sub-second part.
\begin{figure}[t]
    \includegraphics[width=\columnwidth]{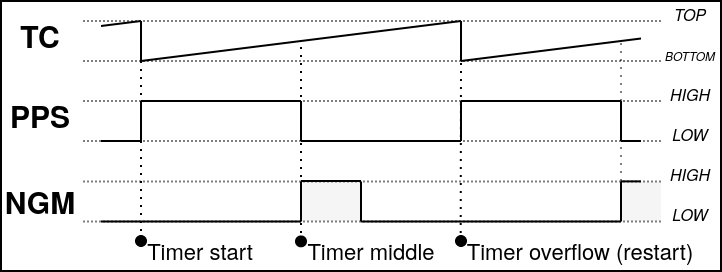}
    \caption{Temporal diagram relating the timer counter values (TC) with the PPS signal and NG messages. The timer performs as (i) real-time clock (RTC), (ii) PPS signal generator, (iii) trigger for NG message transmission start. Timer start corresponds to a new second of RTC and the rising edge of the PPS signal, timer middle corresponds to the falling PPS signal edge and start of NG message transmission.}
\label{fig_tc}
\end{figure}
Another interrupt is triggered by the IMU indicating the readiness of a new inertial data sample. At this instant of time, the MCU saves the current timestamp and sets up the IMU\_DAT\_RDY flag. After gathering the data sample in the main loop, the timestamp along with the data sample is sent to a laptop. Interrupts functionality is crucial for providing high accuracy and precision of LiDAR-IMU time sync. They minimize possible MCU operational delays thanks to utilization of minimum CPU time of the MCU and mainly using the peripheral. Two types of interrupts, described below, do not interfere due to the higher priority of the IMU interrupt.

The main loop stage is sequentially checking two conditions: if a new IMU sample is ready (IMU\_DAT\_RDY flag is set) and if a new NG message needs to be generated and sent to LiDAR (NMEA\_MES\_GEN flag is set). If the first condition is fulfilled, then IMU data are read and sent to the laptop. The NG message containing the current timestamp is generated and sent to LiDAR if the second condition is met. The flags are set at the according interrupt handles (see Fig.~\ref{fig_algo}) and reset in main loop.
The MCU firmware is developed by STM32CubeMX IDE. The ROS driver for reading data is also developed and publicly available on the project page.

\section{Synchronization Performance Evaluation}\label{sec_results}

In this section, we compare the precision of the three methods of time sync: (i) pure synchronization by ROS software, (ii) our improvement based on internal LiDAR timestamping scheme, (iii) external HW timestamping of LiDAR by MCU-platform. We also provide a theoretical estimation of synchronization accuracy of our system.

\begin{figure}[t]
\includegraphics[width=\columnwidth]{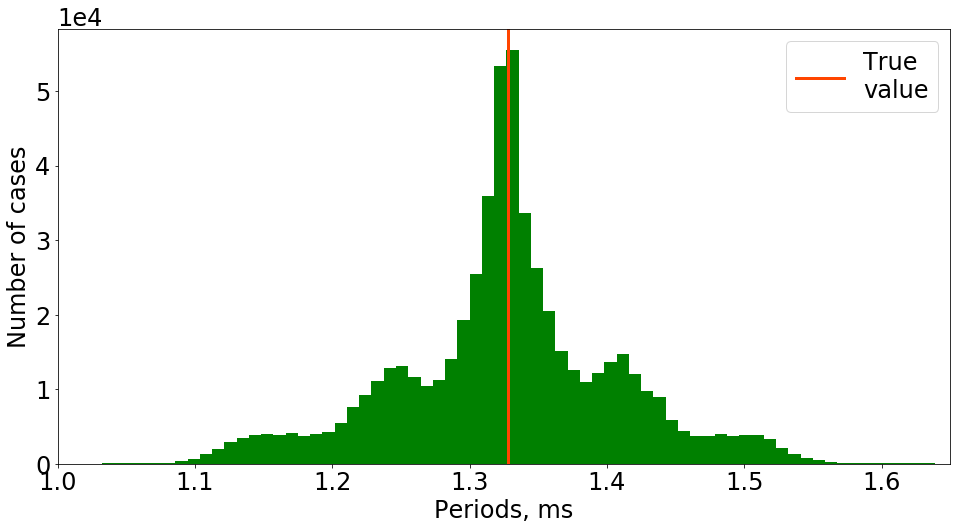}
\caption{Distribution of periods calculated from timestamps given by the pure ROS package~\cite{velodyneros}. The red vertical line expresses the true value of periods (1.328 ms according to~\cite{velodyne}).}
\label{fig_no_GNSS}
\end{figure}

We found that the default LiDAR ROS package~\cite{velodyne} suffers from fluctuations in the timestamping procedure. This happens because the timestamping scheme by default assigns the time of UDP-packets arrival to a computer. These arrivals are loosely related to true time instances of data sampling. We measured the stability of the packet-to-packet period (difference between neighboring timestamps) of LiDAR data for about ten minutes, gathering more than half a million UDP-packets and their timestamps generated by ROS. The histogram of the distribution of periods calculated by these timestamps is depicted in Fig.~\ref{fig_no_GNSS}. We used standard deviation (STD) as a measure of obtained periods' precision. In addition, we measured the difference between the maximum and the minimum obtained period. These results are shown in Table~\ref{tab_results}. 

\begin{table}[t]
    \caption{The timestamping techniques precision comparison}
    \centering
    \begin{tabular}{c c c c}
        \begin{tabular}{@{}c@{}} \textbf{Timestamping}\\ \textbf{technique}\end{tabular} & 
        \begin{tabular}{@{}c@{}} \textbf{STD}\\ \textbf{[\si{\micro\second}]}\end{tabular} & 
        \begin{tabular}{@{}c@{}} \textbf{Min. (max.)}\\ \textbf{period [\si{\micro\second}]}\end{tabular} & 
        \begin{tabular}{@{}c@{}} \textbf{Time sync}\\ \textbf{\textbf{availability}}\end{tabular}\\
        \hline
        ROS & 82.64 & 5.72 (8931.40) & \textbf{Yes}\\
        Internal LiDAR clock & \textbf{0.31} & \textbf{1327.0 (1328.0)} & No\\
        MCU-based clock (ours) & 1.35 & 1327.0 (1365.0) & \textbf{Yes}\\
        [1ex] 
        \end{tabular}
    \label{tab_results}
\end{table}

STD of several decades of microseconds may negatively influence on precision and performance of LiDAR-based navigation and mapping algorithms. For instance, any point in a distance of ten meters from LiDAR will inherit uncertainty of five centimeters in the tangential direction of LiDAR rays (for default LiDAR spinning velocity of ten rounds per second). Outliers (the third row of the table) also harm data processing adding errors.

To eliminate this drawback, we changed the  timestamping scheme by our patch aimed to utilize the internal clock of the LiDAR as a time source for its measurements. Please refer to our project page for a description of the patch~\cite{lidarimusync}. The LiDAR clock starts during the power-on sequence of the LiDAR and is related neither to another sensor time nor global time. According to the table, this method of timestamping has excellent precision comparing the technique described above. However, the key problem of the LiDAR internal clock consists of isolation of LiDAR timestamps from other sensors in a common sensor network (IMU in our case). This issue needs to be solved for correct data fusion of sensors including LiDAR. 

Our system, on the one hand, resolves inter-sensor synchronization; on the other hand, it keeps precision of timestamping at the microsecond order (see Table~\ref{tab_results}).
Precision drops a little because of MCU clock to LiDAR time drift phenomena in-between neighboring PPS signals~\cite{schmid2008exploiting}. 
This drift has a local effect and is not being integrated over time because each subsequent PPS reloads internal LiDAR time values. This fact was checked empirically during the development of the system. LiDAR can be synchronized once by sending a single PPS signal followed by a single NGM but, in this case, time drift will affect synchronization performance by adding an offset that grows over time~\cite{faizullin2021twist}, and we do not follow this strategy.
IMU data timestamping precision is kept at several nanoseconds, which is on the floor of RTC resolution and is not considered in whole precision estimation due to insignificant value of fluctuations.

In this paragraph we provide an estimation of system accuracy according to the sensors' documentation. Empirical estimation involves data driven evaluation methods and is beyond the scope of this work. 
The LiDAR documentation~\cite{velodyne} states no delay between GNSS-based clock and computed data timestamps and provides 1 \si{\micro\second} resolution of data timestamps. 
The IMU delay between a data sample and its corresponding IMU\_DAT\_RDY interrupt is explicitly provided for different setups of the IMU internal low-pass filter with 10 \si{\micro\second} precision~\cite{imu}. All the MCU operating delays related to timestamping are kept under 1 \si{\micro\second} order by design and confirmed in our previous work~\cite{faizullin2021twist}. Considering this findings, we state that time synchronization accuracy is kept as best as 10 \si{\micro\second}. In case of utilization of two or more of the same type LiDARs, the accuracy can reach 1 \si{\micro\second} order.


\section{Conclusion}

In this work, we have introduced a LiDAR time synchronization system to a general spectrum of sensors, including IMU, cameras, other LiDARs, etc. 
Our MCU-based synchronization allows sensor networks to be tightly synchronized at the HW level. Additional data timestamping, as by ROS, is not needed due to automatic online LiDAR timestamps assignment.
The system can work as a tight complement to ROS software as well as be completely independent on it for specific projects.

We have evaluated our system in comparison with SW synchronization via ROS timestamping, showing the superiority in precision of our approach while we maintain all the advantages and flexibility that ROS offers.
We also provided an analysis of precision and showed that our system keeps the same magnitude of precision (1 \si{\micro\second} of STD) as the best possible original clocking schemes of the sensors used. We showed that the synchronization accuracy order is 10 \si{\micro\second} for our setup and can be improved in case of the same type sensors used.


\bibliographystyle{bib/IEEEtran}
\balance
\bibliography{bib/IEEEabrv,bib/bib}

\end{document}